\documentclass[10pt, a4paper]{article}

\usepackage[final]{lrec2026} 

\usepackage{booktabs}
\usepackage{amssymb}

\newcolumntype{W}[1]{>{\hsize=#1\hsize\raggedleft\arraybackslash}X}
\newcolumntype{Y}[1]{>{\hsize=#1\hsize\raggedright\arraybackslash}X}

\title{BenCSSmark: Making the Social Sciences Count in LLM Research}

\name{
Arnault Chatelain$^{1}$,
Étienne Ollion$^{1}$,
Qianwen Guan$^{2}$,
Diandra Fabre$^{3}$, \\\large \bfseries
Lorraine Goeuriot$^{3}$,
Emile Chapuis${^4}$,
Abdelkrim Beloued${^4}$, \\\large \bfseries
Marie Candito$^{2}$,
Nicolas Hervé$^{4}$,
Didier Schwab$^{3}$
}

\address{ 
$^{1}$ \small \itshape CREST (École Polytechnique, ENSAE, CNRS), 5 avenue Le Chatelier, 91120 Palaiseau, France \\
$^{2}$ \small \itshape LLF (Université Paris Cité and CNRS), UFRL Olympe de Gouges, 13 place Paul Ricoeur, 75013 Paris, France \\
$^{3}$ \small \itshape Univ. Grenoble Alpes, CNRS, Grenoble INP, LIG, 38000 Grenoble, France \\
$^{4}$ \small \itshape INA (Institut National de l'Audiovisuel), 4 Avenue de l'Europe, 94366 Bry-sur-Marne, France \\ \small
arnault.chatelain@ensae.fr,
etienne.ollion@polytechnique.edu, \\ \small
\{qianwen.guan, marie.candito\}@u-paris.fr \\ \small 
\{echapuis, abeloued, nherve\}@ina.fr, \\ \small
\{diandra.fabre, lorraine.goeuriot, didier.schwab\}@univ-grenoble-alpes.fr
}

\abstract{
This position paper argues that the under-representation of social science tasks in contemporary LLM benchmarks limits advances in both LLM evaluation and social scientific inquiry. Benchmarks — standardized tools for assessing computational systems — are pivotal in the development of artificial intelligence (AI), including large language models (LLMs). Benchmarks do more than measure progress — they actively structure it, shaping reputations, research agendas, and commercial outcomes. Despite this central role, the social sciences are largely absent from mainstream evaluation frameworks, even though scholars in these fields generate dozens of rigorously annotated, context-sensitive datasets each year. Integrating this work into benchmark design could significantly improve the generalization and robustness of AI models. In turn, models trained on social scientific tasks would likely yield better performance on classic and contemporary tasks in disciplines as diverse as history, sociology, political science or economics. This is all the more pressing as these disciplines are quickly turning to LLMs for assistance. To address this gap, we introduce BenCSSmark, a benchmark composed of datasets annotated by computational social scientists. By integrating social scientific perspectives into benchmarking, BenCSSmark seeks to promote more robust, transparent, and socially relevant AI systems and to foster efficient collaboration.
 \\ \newline  }

\begin{document}

\maketitleabstract

\section{Introduction}
Benchmarks — tools designed to evaluate the performance of computational systems — have played a central role in the development of artificial intelligence (AI) (\citealp{koch2021reduced, raji_ai_2021} ; for a historical perspective, see \citealp{orr_ai_2024}). By providing standardized measures of performance, benchmarks make it possible to compare a system’s quality with that of its predecessors or competitors. When it comes to large language models (LLMs), benchmarking has become an ubiquitous practice, with certain names now familiar to virtually all practitioners in the field. 

Benchmarks are so central that their outcome often determines the success or failure of new models. Today, they shape both the perceived credibility and the market visibility of any LLM. But benchmarks do not merely measure progress; they actively guide it \cite{jaton_constitution_2021}. They steer research priorities toward incremental improvements in benchmark scores \cite{luitse_ai_2025}, and they delineate the scope of what counts as success in the field \cite{crawford_excavating_2021}. Their influence has become so pervasive that some scholars now refer to them as a "lottery” \cite{dehghani_benchmark_2021}, underscoring the extent to which research trajectories and reputations hinge on performance within a specific evaluative framework.

As their importance grew, the number of available benchmarks has dramatically increased (for an extensive review, see \citealp{ni_survey_2025} and \citealp{liu2024datasetslargelanguagemodels}, Section 5). Benchmarks are now used to evaluate and select large language models on a wide variety of tasks and capabilities. Many focus on traditional linguistic tasks in natural language processing, such as paraphrasing \citep{zhang-etal-2019-paws, yang-etal-2019-paws}, named entity recognition \citep{mayhew-etal-2024-universal}, summarization \citep{hasan-etal-2021-xl}, or translation \citep{nllbteam2022languageleftbehindscaling} with some aggregating tasks (e.g. natural language inference, coreference resolution, disambiguation) into general language understanding benchmarks \citep{wang-2018-glue, wang_superglue_2019}. Increasingly, benchmarks have also started to assess models’ reasoning, their general knowledge, their factual consistency, and their alignment with human values \citep{hendrycks_measuring_2020, hendrycks2021aligning, zhao-etal-2024-worldvaluesbench}, sometimes as part of a multitask effort testing several capabilities at the same time \citep{srivastava_beyond_2023, suzgun_challenging_2022}. 

Beyond these general-purpose evaluations, increasingly specialized benchmarks have appeared. Some test mathematical reasoning (e.g., GSM8K, \citep{cobbe_training_2021}, MATH \citep{hendrycks2021measuring}), while others assess code generation and program synthesis (e.g., HumanEval \citep{chen_evaluating_2021}, MBPP \citep{austin_program_2021}). Domain-specific benchmarks now exist across diverse fields, including medicine (e.g., PubMedQA, MedMCQA), law (e.g., LegalBench, CaseHOLD), finance (e.g., FinQA, FPB), and education (e.g., EduBench). This proliferation reflects both the diversification of model applications and the desire to capture domain-sensitive competencies that generic benchmarks often overlook.

Despite this growing diversity, very few benchmarks are specifically designed for tasks relevant to the social sciences. This set of disciplines which includes, among others sociology, economics, political science, history, geography, anthropology, and demography, is also rarely represented in general benchmarks reviews. Popular leaderboards (e.g. Open LLM Leaderboard \citep{huggingface_open_llm_leaderboard}, MTEB Leaderboard \citep{muennighoff-2023-mteb}) only partially incorporate tasks explicitly labeled as social science tasks. And when they are included, this representation is limited.

This absence is unfortunate for two reasons. First, LLMs are increasingly used in the social sciences, and they are often produced with detailed guidelines and test sets. The quality of the annotation is usually excellent, as the tasks tend to be annotated by domain experts, rather than by standard annotators whose work can be less consistent even in mainstream benchmarks \citep{klie-2019-analyzing}. Second, the tasks in the social sciences are original, and they embed cultural differences and diverse viewpoints that contemporary benchmarks lack. 

Yet these available, high-quality resources often remain invisible — uncatalogued, dispersed, and therefore effectively non-existent from a computer scientist's perspective. Because they are not consolidated into available datasets, they do not appear in mainstream suites. Models are subsequently seldom evaluated on them. This probably explains why even state-of-the-art models struggle to meet the needs of social science research, as evidenced by the wide variance in performance on ostensibly similar tasks \citep{ollion2023chatgpt}.

As part of \href{https://pantagruel.imag.fr/}{Pantagruel}, a scientific project dedicated to producing reliable and multimodal large language models (LLMs) in French, we collected various social scientific datasets to evaluate models across various tasks. To this effect, we created \mbox{\textbf{BenCSSmark}}, a benchmark composed of tasks \textit{from actual computational social science projects}. Bringing together scholars from computer science and the social sciences, BenCSSmark pursues two complementary objectives. First, it aims to encourage NLP scholars to discover the richness of available datasets in the social sciences. Second, it seeks to guide model development by focusing on tasks pertaining to domains that are understudied. Our hope is indeed that this initiative will improve the quality of future models for social scientific tasks, and overall. It is also that \mbox{BenCSSmark's} example will highlight the relevance of social science tasks and lead to a better integration into model evaluations.

\section{Benchmarks and Social Sciences: A Limited Interaction}
\subsection{Related Works}

The encounter between AI and the social sciences has proven both rich and productive. On the one hand, social scientists have devoted considerable attention to AI as a cultural and social phenomenon in its own right. They have examined its diffusion, the forms of resistance it generates, and the social, economic, and institutional transformations it brings about \citep{tubaro2020trainer, acemoglu2021harms, bryan2026econimpact}. They have also examined how AI systems — trained on vast corpora of human-generated data — reproduce, reinterpret, and at times amplify the social norms, biases, and imaginaries embedded in the societies from which these data originate \citep{garg2018word, bender2021dangers}.

In addition to these social sciences of AI, researchers have also explored the use of AI in the social sciences, following the distinction proposed by \citet{xu_ai_2024}. This includes earlier techniques derived from machine learning \citep{athey2019machine}, as well as more recent advances in generative AI \citep{bail_can_2024} and, in particular, large language models \citep{ziems_can_2024}. Across the social sciences, these tools have inspired new research practices, sparked methodological debates, and fueled growing controversies regarding their epistemological implications and limits \citep{binz_how_2025, boelaert2025machine}.

However, despite these fruitful intersections, the development of benchmarks either derived from or tailored to the social sciences has lagged. It is telling that a recent and comprehensive benchmark survey does not include many tasks from these disciplines, even in the section devoted to the social sciences \citep{ni_survey_2025}. 

This initial observation does not imply that the social sciences are altogether absent from the natural language processing (NLP) literature on evaluation. Recently, two projects have proposed such benchmarks in adjacent directions. One such initiative is HSSBench \citep{kang_hssbench_2025}, an extensive benchmark designed to test multimodal models on tasks related to human knowledge. It presents several thousand questions in six different languages, requiring a model to combine advanced factual knowledge with visual recognition. However, its primary orientation is toward evaluating visual or multimodal models, not models aimed specifically at social science questions.

Another important initiative, in part closer to our own, is the work initiated by \citet{li_social_2024}. The authors aggregated 480 datasets designed to measure “social intelligence.” Their tasks are designed to assess a model’s capacity to interpret cognitive, situational, and behavioral cues in interactional contexts. While some of these tasks (such as blame attribution, irony or sarcasm detection, nuanced abuse classification, and toxicity moderation) could indeed be relevant to social science research, Li et al.’s benchmark differs in emphasis: it primarily targets dialogue-based understanding, often to improve conversational agents rather than the broader tasks of empirical social inquiry.

Though seemingly more distant, other projects also merit mention. A recent review, for instance, compiled studies that account for the diversity of viewpoints expressed on a given issue and assembled datasets in which multiple, and sometimes conflicting, annotations are preserved \citep{frenda_perspectivist_2025}. This perspectivist approach, which challenges the notion of a unique ground truth or gold standard, draws heavily on insights from the social sciences — and in particular on the idea that evaluative judgments are themselves situated and shaped by experience. As \citet{rottger_two_2022} have argued, annotation can follow different epistemological logics: some approaches aim to resolve disagreement and converge toward consensus, while others deliberately retain divergence to reflect the plurality of perspectives embedded in language.

\subsection{Making Sense of the Absence}

Why are social science benchmarks so scarce when it comes to LLM evaluation? Several factors help explain this. First, unlike in machine learning, these disciplines have not historically relied on benchmarks as tools for the cumulative improvement of knowledge. This explains the paradox whereby recent articles systematically report performance metrics or quality assessment, yet no centralized compilation of these evaluations can be found. Assessments are often carried out within individual studies — sometimes using large, publicly available datasets \cite{ziems_can_2024} or \textit{ad hoc} collections created for a specific project (e.g., \citealt{gilardi_chatgpt_2023}). But these datasets are rarely standardized, shared or made available for systematic testing. The absence of consistent data formats and shared repositories thus makes the creation of reproducible benchmarks particularly labor-intensive.

Another reason lies in the concentration of current NLP research on social issues over a limited set of tasks, those that most closely resemble traditional NLP applications. Sentiment analysis, hate speech or toxicity detection, misinformation detection, and stance detection are among the best-developed \cite{thapa_large_2025}, in large part because they have clear commercial applications that attract funding. By contrast, few social science tasks can be directly framed as commercially relevant. Yet existing benchmarks only represent a fraction of the methodological and analytical needs of the social sciences, whether in text classification, information extraction, or corpus-based analysis. And even within these relatively standardized tasks, persistent difficulties remain. Contextual variation — linguistic, cultural, or situational — often challenges model generalization. For instance, \citet{nogara_toxic_2025} demonstrate that Perspective API — often considered state-of-the-art in online hate detection — systematically assigns higher toxicity scores to German-language content than to comparable English inputs, highlighting the uneven cross-linguistic and contextual robustness of contemporary AI systems.

A third difficulty lies in the fact that social science tasks are rarely standardized. On the contrary, research in these disciplines often relies on context-specific categories and tailor-made analytical frameworks. This is partly due to the recurrent debates about the definition of key phenomena (e.g. "social class" or "populism") and disagreements on how best to measure such concepts, which often lead to different operationalizations. More importantly, this is also because social science research questions often require bespoke tools. For instance, in a study of interactions between elected officials and citizens, \citet{claesson2025} sought to measure the proportion of messages received by politicians that were not hateful or toxic, but also simply "critical", and even "supportive". As no existing tool could accurately capture this distinction, she developed her own classifier. While this poses clear challenges from a knowledge cumulation perspective, it also represents a condition for the inventiveness and interpretive nuance that define these disciplines.

\subsection{Social Science Tasks}

The preceding discussion points to a broader lack of conceptual clarity regarding what constitutes a social science task — an ambiguity that may have further obscured their absence from widely used benchmarks. In our view, social science tasks are those that exhibit construct validity with respect to social scientific concepts or questions. This definition implies not only that the task has been employed within the social sciences, but also that it has been regarded as a first-best operationalization of a given approach. In most cases, such tasks have required researchers to produce \emph{ad hoc} expert annotations. However, these annotations are rarely reused or incorporated into existing benchmarks.

It could be argued that most of these tasks would still fall into already existing well-known NLP tasks categories, representing only marginal variations. We contend that their value lies precisely in introducing these variations and nuances.

\section{BenCSSmark}

\subsection{The Initiative}
BenCSSmark was created as part of Pantagruel, a collective research project aiming to develop the next generation of large language models (LLMs) in French. The interdisciplinary team includes researchers from the social sciences, who were not brought into the project to focus on the social implications of AI or to audit bias, but rather to integrate tasks directly derived from their own disciplines. Their contribution centers on designing and curating a set of tasks specific to research in the social sciences. The underlying motivation is that as these disciplines increasingly engage with AI tools, they generate datasets that can improve the linguistic and analytical quality of language models.

\subsection{Data Collection Strategies}
We pursued two strategies simultaneously: the first consisted of producing entirely new data. Common in NLP, this approach involves assembling a team of annotators, training them for a specific task, and generating a dataset. In our case, all annotations were preserved — in addition to the adjudicated decision — in order to reflect the diversity of viewpoints that can emerge around a given task.

The second method, more typical of the social sciences, relies on datasets annotated in great detail by a single individual, often an expert in the relevant domain \cite{do-ollion-2022}. Such data cannot be analyzed using conventional procedures, such as inter-annotator agreement metrics. Nevertheless, these annotations are numerous. They are also annotated by experts, and they span a wide range of topics, areas and period. As a result, they constitute an invaluable source for constructing benchmarks that can help develop models better attuned to the interpretive and contextual needs of social scientific research.

\subsection{Three Principles for a Benchmark}

BenCSSmark was created with three guiding principles in mind aimed to maximize its relevance. The first principle is to better represent the diversity of viewpoints and approaches of social science disciplines. To do so, we draw tasks from underrepresented fields such as sociology and political science. More importantly, for more classical NLP problems such as text classification tasks such as topic, frame or stance detection, we made sure to respect how these disciplines have been approaching and discussing these tasks. To be sure, a social scientist would hardly consider frame detection on social media posts today to be the same task as frame detection on newspaper articles from the first half of the 20th century — at least progress on the former would not be considered relevant for applications on the latter. We also included more unusual tasks originating from these disciplines such as concept detection (searching for instances of populism, of gender, etc.).

The second principle guiding our benchmark is to reflect contextual, socio-cultural and temporal variations. Such variations lie at the heart of social scientific inquiry, which seeks to uncover both enduring regularities and significant shifts in how societies classify situations and name phenomena. Our effort, therefore, consists in assembling data from different data genres and historical periods in order to cover as wide a range of topics, populations, and contexts as possible.

The final principle guiding our work is to preserve, whenever possible, the diversity of annotators’ perspectives. For all tasks that we annotated ourselves, we therefore produced two versions of the data: one with a single ground truth (as is commonly done) and one where all single annotations have been kept. These data presents two key advantages. First, it allows us to evaluate task difficulty in a conventional manner — low agreement among annotators may indicate greater ambiguity or complexity, while high agreement suggests clearer boundaries and greater reliability. Second, it contributes to the perspectivist call for producing disaggregated data\footnote{\url{https://pdai.info/}}.

For these reasons, our datasets are enriched with relevant metadata. This includes information about the task and disciplinary context, as well as detailed information on the text data  and the time period covered. Whenever feasible, we also collect metadata about annotators, enabling their annotations to be situated within their social and contextual backgrounds — a practice that a small number of datasets in the perspectivist literature have begun to adopt (\citealp{frenda_perspectivist_2025}, p. 1713).

\subsection{Data}

A crucial aspect of our initiative is the systematic categorization of datasets across multiple facets. Each task is first assigned to a broad task category. The aim here is to characterize the literature related to the task in NLP. Tasks are then also categorized according to the concepts the researchers have been looking to capture. This concept categorization is novel and is introduced to make explicit the specificity of each of these \textit{social science} tasks. Together these two taxonomies look to further exchanges between computer and social scientists. Additional information focuses on the dataset characteristics (data genre, descriptive statistics, temporal scope, modality). This multidimensional organization allows for more precise selection and comparison across contexts.

At the time of writing, BenCSSmark contains 27 datasets, all in French (see \autoref{tab:collected_data}). This number is expected to grow as the project advances and we increase the range of social scientists we reach out to. Beyond presenting our current work, this article aims to invite researchers in the social sciences and natural language processing to contribute to a similar initiative.

Most of the collected tasks fall into common and well-studied task categories such as topic classification, frame detection, hate-speech detection, bias detection, quote detection, or coreference resolution. However, they often deviate from the typical use cases of these categories in terms of the concepts studied and/or the data to which they are applied. Examples include topic classification of music-related articles in the arts \& culture sections of the national press; topic-specific frame detection with politically-loaded angles\footnote{For instance, discussing taxation-related topics presenting it as an undue pressure on the wealthy}; distinguishing hateful from critical comments in political tweets;  detecting biased statements on Wikipedia; detecting unattributed quotes in the press or using a coreference resolution task to measure how much newspapers cite sources with different political leaning.

The benchmark also includes task categories that remain comparatively less formalized, such as concept detection or argumentative strategy detection.  For the former the tasks we collected focus on detecting gender-related or social class-related research papers in a corpus of social science paper abstracts. For the latter, the tasks correspond to detecting when opinion pieces on the radio or in politicians' discourses push their arguments using certain types of persuasion strategies.  

The remaining task category (``other detection'') groups together \textit{ad hoc} tasks which did not fit in any of the common task categories. The concepts these tasks study illustrate how varied and creative social science tasks can be. A notable consequence of the idiosyncrasy of these tasks is that model performance on them is unknown -- thus providing novel tests to assess the generalization capacities of models. Examples include detecting when newswriting on music-related topics is prescriptive (e.g. album reviews, tour announcement, interviews), detecting when politicians make reform pledges in their electoral manifestos, or detecting when opinion pieces on the radio make political prophecies. 

Regarding the tasks' types, BenCSSmark features binary and multiclass classification, multilabel classification, span detection and coreference resolution. As the benchmark expands we hope to include other types of tasks that may be relevant to the social sciences, such as semantic text similarity or clustering. In doing so we also hope to popularize these other approaches among social scientists. 

The text genres cover social media, news, political discourses, broadcast discussions, broadcast news, electoral manifestos and Wikipedia and social science articles. In terms of time period, the datasets currently span the second half of the 20th century onwards. Earlier datasets pose distinctive challenges — archaic vocabulary, shifting social categories, and changes in discourse conventions — that make them particularly valuable for evaluating the robustness and adaptability of modern language models. As for the units of annotation, BenCSSmark contains examples of both sentence, paragraph and full text annotations. The gender concept detection tasks (tasks number 6 and 7) notably produced annotations at both the sentence and the paragraph-level, allowing model performance on each to be compared.

The project currently covers two modalities: written text and speech transcription\footnote{It is the case for the radio and TV data.}. Speech transcription poses distinct challenges, notably residual transcription errors — despite substantial recent progress in automatic speech recognition — as well as the complexity of multi-speaker dialogues, which necessitate dedicated formatting and preprocessing procedures. 

\begin{table*}[ht!]
    \centering
    \scriptsize
    \begin{tabularx}{\linewidth}{@{}W{0.3}Y{1.4}Y{1.3}Y{2.2} Y{1.3}Y{1.8}Y{1.3}W{0.75}@{\phantom{a}}Y{0.85} W{0.25}W{0.25}W{0.2}@{}}
    \toprule
     &\multicolumn{3}{l}{\textbf{Task}} & \multicolumn{5}{l}{\textbf{Data}} & \multicolumn{3}{l}{\textbf{Annotators}}\\
     \textbf{\#} & \textbf{Category} & \textbf{Type} & \textbf{Concept} & \textbf{Genre} &  \textbf{Description} & \textbf{Period} & \textbf{Size} & \textbf{Unit} & \textbf{XP} & \textbf{RA} & \textbf{P} \\
    \midrule
1 & Argumentative strategy detection & binary classif. & Appeal to authority & radio & opinion pieces & 2017-2023 & 900 & parag. &  & 4 & \checkmark \\\addlinespace[3pt]
2 &  &  & Appeal to majority &  &  &  & 900 & parag. &  & 4 & \checkmark \\\addlinespace[3pt]
3 &  &  & Appeal to majority & speech & politicians' speeches & 1974-2022 & 900 & parag. &  & 4 & \checkmark \\\midrule[0.1pt]\addlinespace[3pt]
4 & Bias detection & binary classif. & Non-neutral statement & wikipedia pages & politicians wikis & 2002-2024 & 130 & varied & 5 &  & \checkmark \\\midrule[0.1pt] \addlinespace[3pt]
5 & Bias labeling & 3-label classif. & Non-neutral statement bias type & wikipedia pages & politicians wikis & 2002-2024 & 130 & varied & 1 &  &  \\\midrule[0.1pt] \addlinespace[3pt]
6 & Concept detection & binary classif. & Gender & academic articles & social science abstracts & 2001-2025 & 2,889 & parag. & 1 &  &  \\\addlinespace[3pt]
7 &  &  & Gender &  &  &  & 4,100 & sentence & 2 &  &  \\\addlinespace[3pt]
8 &  &  & Social class &  &  &  & 1,990 & parag. & 2 &  &  \\\midrule[0.1pt] \addlinespace[3pt]
9 & Coreference resolution & pairwise coref. & Press political source intensity & press & national dailies & 1998-2020 & 200 & full text & 2 & 5 &  \\\midrule[0.1pt] \addlinespace[3pt]
10 & Frame detection & 3-class classif. & Immigration framing & press & national dailies & 2000-2019 & 1,601 & parag. & 1 &  &  \\\addlinespace[3pt]
11 &  &  & LGBT rights framing &  &  &  & 2,996 & parag. & 1 &  &  \\\addlinespace[3pt]
12 &  &  & Taxation framing &  &  &  & 2,144 & parag. & 1 &  &  \\\midrule[0.1pt] \addlinespace[3pt]
13 & Hate speech detection & binary classif. & Abusive comment & tweets & politics-related tweets & 2022-2023 & 2,152 & full text & 1 &  &  \\\addlinespace[3pt]
14 &  &  & Critical comment &  &  &  & 651 & full text & 1 &  &  \\\midrule[0.1pt] \addlinespace[3pt]
15 & Other detection & binary classif. & Inclusive language & academic articles & social science abstracts & 2001-2025 & 1,511 & parag. & 1 &  &  \\\addlinespace[3pt]
16 &  &  & Reform pledge & electoral manifestos & French party programs & 2015-2024 & 28,215 & sentence & 1 &  &  \\\addlinespace[3pt]
17 &  &  & Political forecasting & radio & opinion pieces & 2017-2023 & 1,650 & parag. &  & 4 & \checkmark \\\addlinespace[3pt]
18 &  &  & Music newswriting (prescription detection) & press & national dailies & 1998-2023 & 1,450 & full text & 1 &  &  \\\addlinespace[3pt]
19 &  &  & Supportive comment & tweets & politics-related tweets & 2022-2023 & 1,174 & full text & 1 &  &  \\\midrule[0.1pt] \addlinespace[3pt]
20 & Quote detection & span detect. & Press political source diversity & press & national dailies & 1998-2020 & 120 & full text & 2 & 5 &  \\\addlinespace[3pt]
21 &  & binary classif. & Unattributed quote & press & national dailies & 1945-2018 & 10,633 & sentence & 2 & 3 &  \\\addlinespace[3pt]
22 &  & span detect. & Unattributed quote &  &  &  & 10,633 & sentence & 2 & 3 &  \\\midrule[0.1pt] \addlinespace[3pt]
23 & Topic classif. & binary classif. & Music-related content & press & national dailies & 1998-2023 & 1,450 & full text & 1 &  &  \\\addlinespace[3pt]
24 &  & 13-class classif. & News categories & press & national dailies & 1945-2022 & 2,000 & full text &  & 1 &  \\\addlinespace[3pt]
25 &  & 24-class classif. & Policy issues & tweets & politics-related tweets & 2008-2023 & 6,386 & full text & 1 &  &  \\\addlinespace[3pt]
26 &  & 3-class classif.  & Political newswriting (horserace detection) & press & national dailies & 1945-2018 & 3,843 & sentence & 2 & 3 &  \\\midrule[0.1pt] \addlinespace[3pt]
27 & Topic labeling & 115-label classif. & Thematic categorization & radio/tv & radio/tv transcripts & 1982-2025 & 2,500 & parag. & 28 &  &  \\
    \bottomrule
    \end{tabularx}
    \caption{Collected Datasets. \textbf{Category} corresponds to common NLP terminology for tasks. \textbf{Concept} focuses on what social scientists have been trying to capture. \textbf{XP} stands for Experts, \textbf{RA} for Research Assistants and \textbf{P} stands for Perspectivism and indicates whether individual annotations are available. All tasks come from ongoing or published social science projects. See \autoref{tab:table_sources} in the appendix for tasks' authors.}
    \label{tab:collected_data}
\end{table*}

\subsection{The Social Sciences as a Stress Test for Artificial Intelligence}

When building a dataset for the social sciences, we are not only intent on improving models performances for social scientific applications. Our aim is also to leverage the richness of their tasks to hopefully advance research on large language models more broadly.

Large language models are often evaluated on tasks that prioritize precision, factual recall, or narrow reasoning abilities. Yet these metrics fail to capture one of the most demanding dimensions of human intelligence and diversity: the capacity to interpret meaning in context, to navigate ambiguity, and to arbitrate between conflicting perspectives. The perspectivist literature has started to point out such limits but translating these insights into evaluation metrics remains an open question \citep[one exception being][]{gordon_2021_disagreement}.

Social science tasks offer precisely this kind of challenge. By confronting models with historically and culturally situated data, multiple viewpoints, and conceptually fluid categories, they expose weaknesses that remain invisible in conventional benchmarks. They require models to handle pragmatic nuance, shifting semantics, and perspectival disagreement — dimensions that are central to human communication but peripheral to most NLP evaluations, particularly when it comes to annotation practices.

From this standpoint, BenCSSmark can be viewed as a stress test for AI. It evaluates not only linguistic competence but also interpretive robustness and socio-cognitive flexibility. Performance on social science tasks thus becomes an indicator of a model’s broader generalization capacity — its ability to deal with variability, inconsistency, and moral ambiguity inherent in human data. Additionally BenCSSmark also serves as a call for developing novel evaluation metrics that better account for subjectivity or ambiguity in model scoring.

\section{Limitations and Future Work}
\subsection{A First Step Only}\label{sec:first_step}

This initiative naturally comes with limitations. The first concerns the current scale of the data, which remains limited in both quantity and scope. It is  restricted to a limited range of media and forms of expression. Another limitation of BenCSSmark lies in the current nature of the data. At the moment, due to the goals of the project, the datasets are exclusively textual, and only in French. The media covered remain largely confined to conventional formats (such as newspapers and official speeches) and well-studied domains (such as social media posts). 

Yet this initiative should first be read as an invitation: an open call for social scientists to engage with and contribute to benchmarks, in the spirit of a collaborative project designed to serve the collective advancement of both NLP and the social sciences. Our aim with this project is thus not to build a definitive benchmark, but rather to draw the attention of both NLP practitioners and social science researchers to the importance of more sustained exchange between the two communities. 

NLP researchers could benefit from training, selecting, and validating their models on the numerous and fine-grained data that already exist within the social sciences, which often simply lack visibility. Because they pertain to fundamentally human activities involving language, of central interest to linguists (and to models), and because they introduce diversity and nuance into computational tasks, social science data constitute a valuable reservoir for NLP. Conversely, for social scientists, having access to models trained on tasks aligned with their research needs would clearly be advantageous. By identifying the specific needs of these disciplines — which constitute a substantial share of contemporary research — we can help ensure that they are adequately represented and integrated into the development of future models and benchmarks.

Another limitation is that BenCSSmark remains closed, at least for the time being. Our initial objective was to provide a fully open repository. However, we were compelled to revise this plan, as much of the data is either proprietary (e.g., newspaper archives) or may contain personally identifiable information. In both cases, hosting the data within a shared infrastructure would have exposed us to legal and ethical liabilities, and we therefore decided against this option for the time being.

Moreover, as large language model (LLM) developers increasingly incorporate existing datasets into their training corpora, benchmarks risk becoming rapidly outdated — a challenge that currently affects many widely used evaluation frameworks. These issues are not specific to our project, and addressing them will require the development of innovative infrastructures and governance models in the future.

\subsection{Avoiding the Risks of Benchmarkisation}

Another risk inherent in this initiative is that it may reproduce the very shortcomings commonly associated with benchmarking — particularly the distortions that arise from the excessive reliance on standardized evaluation metrics. These limitations have been extensively documented in the literature.

Benchmarks, by construction, tend to narrow the definition of what counts as relevant to what can be measured. They may foster optimization toward the metric rather than toward the phenomenon of interest (Goodhart’s Law), leading models to overfit to test sets instead of improving in generalizable ways. Moreover, when benchmarks become central instruments of evaluation, they risk disciplining research agendas, privileging technical improvement over theoretical or conceptual innovation \citep{hooker1995testing, paullada_data_2021}. 

Benchmarks can thus become unproductive. This fact has been extensively discussed in the NLP community \citep{liao2021are, weidinger2025toward, raji_ai_2021}. It raises acute questions for a set of disciplines as diverse and interpretively rich as the social sciences. Excessive standardization could undermine the pluralism of approaches that characterizes the discipline, silencing context, uncertainty, and disagreement — precisely the elements that constitute its epistemic strength. 

A straightforward way to guard against this drift is to design benchmarks that incorporate multiple labels — such as time period, discipline, cultural area, and task type — thereby enabling evaluations to be filtered according to relevant criteria rather than collapsed into a single aggregate score. Another protection is to continually expand such benchmarks with new datasets, thereby introducing new research questions and perspectives over time.

\section{Conclusion}

Benchmarks have long guided progress in large language model (LLM) research. They focus community efforts on standardized tasks and enable direct comparisons between models. However, tasks that are not included in benchmarks are often neglected: models are not optimized for them and their performance on such tasks remains largely unknown. This has notably been the case for \textit{social science tasks}, despite the wide range of datasets produced by social scientists in their research.

As an initial step toward addressing this gap, we introduced BenCSSmark, a benchmark composed of social science tasks from ongoing social science research projects or published papers. Our objectives are threefold: to help conceptualize social science tasks and highlight their diversity; to provide an initial benchmark tailored to them; and to foster dialogue between computer scientists and social scientists by engaging with their distinct research traditions.

Expanding LLM benchmarking to better represent the social sciences presents opportunities for both communities. For social scientists, it offers a way to shape model development toward tools that better address their research questions. For computer scientists, it opens access to a growing body of carefully annotated datasets. By making the social sciences count, our hope is that we will see improvements in both fields.

\section{Acknowledgements}

This research has been partially funded by the French National Research Agency (ANR project "PANTAGRUEL", ANR-23-IAS1-0001). It is also partly supported by Hi! PARIS and the ANR/France 2030 program (ANR-23-IACL-0005). It also received government funding managed by ANR under France 2030, reference ANR-23-IACL-0006. 

\section{Bibliographical References}\label{sec:reference}

\bibliographystyle{lrec2026-natbib}
\bibliography{references}

\clearpage

\appendix

\begin{table*}[t]
    \parbox{\columnwidth}{
      \section{Additional Table}
    }
    \vspace{1em}
    \centering
    \scriptsize
    \begin{tabularx}{\linewidth}{@{}W{0.2}Y{1.2}Y{1.1}Y{1.3}Y{1}Y{1.1}XY{1.1}@{}} 
        \toprule
        \textbf{\#} & \textbf{Category} & \textbf{Type} & \textbf{Concept} & \textbf{Data} & \textbf{Contact} & \textbf{Status} & \textbf{Reference} \\
\midrule
  1 & Argumentative strategy detection &    binary classif. &                         Appeal to authority &                radio &                            \href{https://www.linkedin.com/in/yacine-chitour-721b02207/}{Yacine Chitour} & ongoing work &                                \\\addlinespace[3pt]
  2 &  &     &                          Appeal to majority &                 &                            \href{https://www.linkedin.com/in/yacine-chitour-721b02207/}{Yacine Chitour} & ongoing work &                                \\\addlinespace[3pt]
  3 &  &     &                          Appeal to majority &               speech &                            \href{https://www.linkedin.com/in/yacine-chitour-721b02207/}{Yacine Chitour} & ongoing work &                                \\\midrule[0.1pt] \addlinespace[3pt]
  4 &                   Bias detection &    binary classif. &                       Non-neutral statement &      wikipedia pages &       \href{https://www.linkedin.com/in/victor-planche-a25687269/?originalSubdomain=fr}{Victor Planche} & ongoing work &                                \\\midrule[0.1pt] \addlinespace[3pt]
  5 &                    Bias labeling &   3-label classif. &             Non-neutral statement bias type &      wikipedia pages &       \href{https://www.linkedin.com/in/victor-planche-a25687269/?originalSubdomain=fr}{Victor Planche} & ongoing work &                                \\\midrule[0.1pt] \addlinespace[3pt]
  6 &                Concept detection &    binary classif. &                                      Gender &    academic articles (parag.) &                   \href{https://scholar.google.com/citations?user=BCnoDlEAAAAJ\&hl=en}{Julien Boelaert} &    published &    \citet{boelaert2025part} \\\addlinespace[3pt]
  7 &                 &     &                                      Gender &  academic articles (sentence)  &                   \href{https://scholar.google.com/citations?user=BCnoDlEAAAAJ\&hl=en}{Julien Boelaert} &    published &    \citet{boelaert2025part} \\\addlinespace[3pt]
  8 &                 &     &                                Social class &    &                   \href{https://scholar.google.com/citations?user=BCnoDlEAAAAJ\&hl=en}{Julien Boelaert} &    published &    \citet{boelaert2025part} \\\midrule[0.1pt] \addlinespace[3pt]
  9 &           Coreference resolution &    pairwise coref. &            Press political source intensity &                press &                                            \href{https://emmabonuttidagostini.github.io/}{Emma Bonutti} & ongoing work &                                \\\midrule[0.1pt] \addlinespace[3pt]
 10 &                  Frame detection &   3-class classif. &                         Immigration framing &                press &                                                       \href{https://rubingshen.github.io/}{Rubing Shen} &  PhD chapter & \citet[chap. 3]{Shen2024}   \\\addlinespace[3pt]
 11 &                  &    &                         LGBT rights framing &                 &                                                       \href{https://rubingshen.github.io/}{Rubing Shen} &  PhD chapter & \citet[chap. 3]{Shen2024}   \\\addlinespace[3pt]
 12 &                   &   &                            Taxation framing &                 &                                                       \href{https://rubingshen.github.io/}{Rubing Shen} &  PhD chapter & \citet[chap. 3]{Shen2024}   \\\midrule[0.1pt] \addlinespace[3pt]
 13 &            Hate speech detection &    binary classif. &                             Abusive comment &               tweets &                                                    \href{https://anninacla.github.io/}{Annina Claesson} &    published &           \citet{claesson2025} \\
 14 &            &     &                            Critical comment &                &                                                    \href{https://anninacla.github.io/}{Annina Claesson} &    published &           \citet{claesson2025} \\\midrule[0.1pt] \addlinespace[3pt]
 15 &                  Other detection &    binary classif. &                          Inclusive language &    academic articles &                   \href{https://scholar.google.com/citations?user=BCnoDlEAAAAJ\&hl=en}{Julien Boelaert} &    published &    \citet{boelaert2025part} \\\addlinespace[3pt]
 16 &                 &     &                 Institutional reform pledge & electoral manifestos &                            \href{https://www.pacte-grenoble.fr/fr/frederic-gonthier}{Frédéric Gonthier} & ongoing work &                                \\\addlinespace[3pt]
 17 &                 &     &  Music newswriting (prescription detection) &                press &                                                      \href{https://scoavoux.github.io/}{Samuel Coavoux} & ongoing work &                                \\\addlinespace[3pt]
 18 &                   &     &                       Political forecasting &                radio &                            \href{https://www.linkedin.com/in/yacine-chitour-721b02207/}{Yacine Chitour} & ongoing work &                                \\\addlinespace[3pt]
 19 &                  &     &                          Supportive comment &               tweets &                                                    \href{https://anninacla.github.io/}{Annina Claesson} &    published &           \citet{claesson2025} \\\midrule[0.1pt] \addlinespace[3pt]
 20 &                  Quote detection &       span detect. &            Press political source diversity &                press &                                            \href{https://emmabonuttidagostini.github.io/}{Emma Bonutti} & ongoing work &                                \\\addlinespace[3pt]
 21 &                   &    binary classif. &                          Unattributed quote &                press &                                                       \href{https://rubingshen.github.io/}{Rubing Shen} &  PhD chapter & \citet[chap. 2]{Shen2024}   \\\addlinespace[3pt]
 22 &                   &       span detect. &                          Unattributed quote &                press &                                                       \href{https://rubingshen.github.io/}{Rubing Shen} &  PhD chapter & \citet[chap. 2]{Shen2024}   \\\midrule[0.1pt] \addlinespace[3pt]
 23 &             Topic classification &    binary classif. &                       Music-related content &                press &                                                      \href{https://scoavoux.github.io/}{Samuel Coavoux} & ongoing work &                                \\\addlinespace[3pt]
 24 &             &  13-class classif. &                             News categories &                press & \href{https://www.linkedin.com/in/francesco-colonna-3aa0581a1/?originalSubdomain=uk}{Francesco Colonna} & ongoing work &                                \\\addlinespace[3pt]
 25 &              &  24-class classif. &                               Policy issues &               tweets &                                                     \href{https://malojan.github.io/website/}{Malo Jan} & ongoing work &                                \\\addlinespace[3pt]
 26 &             &   3-class classif. & Political newswriting (horserace detection) &                press &                                                            \href{https://sally14.github.io/}{Salomé Do} &    published &      \citet{do-ollion-2022}    \\\midrule[0.1pt] \addlinespace[3pt]
 27 &                   Topic labeling & 115-label classif. &                     Thematic categorization &             radio/tv &                                                           \href{https://www.herve.name/}{Nicolas Hervé} & INA internal &                                \\
\bottomrule
    \end{tabularx}
    \caption{Authors and References for the Collected Datasets. \textbf{\#} uniquely identifies each dataset. \textbf{Category} corresponds to common NLP terminology for tasks. \textbf{Concept} focuses on what social scientists have been trying to capture. \textbf{Data} presents data genre and the unit of text annotation in parenthesis when needed for disambiguation. \textbf{Contact} provides a link to one of the data author's websites. \textbf{Status} is the research project status.}
    \label{tab:table_sources}
\end{table*}

\end{document}